\documentclass[
  reprint,
  amsmath,amssymb,
  aps,
  prl,         
  floatfix    
]{revtex4-2}

\usepackage[english]{babel}

\usepackage{graphicx}
\usepackage{dcolumn}
\usepackage{bm}
\usepackage{amsmath}
\usepackage{cases}
\usepackage{color}

\usepackage{color}
\usepackage{hyperref}   

\usepackage{xr}

\makeatletter
\newcommand*{\addFileDependency}[1]{
  \typeout{(#1)}
  \@addtofilelist{#1}
  \IfFileExists{#1}{}{\typeout{No file #1.}}
}
\makeatother

\begin{document}



\title{Collisional Binding and Transport of Shape-Changing Robot Pairs}
\date{\today}

\author{Akash Vardhan $^1$}
\author{Ram Avinery $^{1,2}$}
\author{Hosain Bagheri $^1$}
\author{Velin Kojohourav $^{1,3}$}
\author{Shengkai Li $^{1,4}$}
\author{Hridesh Kedia $^5$}
\author{Tianyu Wang $^{6,7}$}
\author{Daniel Soto $^1$}
\author{Kurt Wiesenfeld $^1$}
\author{Daniel I. Goldman$^{1}$}
\email{Corresponding author: daniel.goldman@physics.gatech.edu}

\affiliation{$^1$School of Physics, Georgia Institute of Technology, Atlanta, GA, USA}
\affiliation{$^2$Apple Inc., Cupertino, CA, USA}
\affiliation{$^3$Department of Mechanical Engineering, Stanford University, Stanford, CA, USA}
\affiliation{$^4$Department of Physics, Princeton University, Princeton, NJ, USA}
\affiliation{$^5$Department of Physics, Bar-Ilan University, Ramat Gan, Israel}
\affiliation{$^6$Institute for Robotics and Intelligent Machines, Georgia Institute of Technology, Atlanta, GA, USA }
\affiliation{$^7$George W. Woodruff School of Mechanical Engineering, Georgia Institute of Technology, Atlanta, GA, USA }

\begin{abstract}

We report in experiment and simulation the spontaneous formation of dynamically bound pairs of shape changing robots undergoing locally repulsive collisions. These physical `gliders' robustly emerge from an ensemble of individually undulating three-link two-motor robots and can remain bound for hundreds of undulations and travel for multiple robot dimensions. Gliders occur in two distinct binding symmetries and form over a wide range of angular oscillation extent. This parameter sets the maximal concavity which influences formation probability and translation characteristics. Analysis of dynamics in simulation reveals the mechanism of effective dynamical attraction -- a result of the emergent interplay of appropriately oriented and timed repulsive interactions. Tactile sensing stabilizes the short-lived conformation via concavity modulation.
\end{abstract}

\maketitle
Collisional dynamics govern interactions across scales, from particle collisions revealing fundamental forces to the scattering of electrons in crystals. While classical inelastic collisions, which do not conserve energy, can lead to emergent phenomena like clustering \cite{Olafsen1998clustering,Aranson2006patterns}, inelastic collapse  \cite{Zhou1996inelastic,Cornell1998inelastic,McNamara1994inelastic,Goldman1998absence}, jamming, and transitions between fluid- and solid-like states \cite{Clerc2008liquid,Behringer2018jamming}, active systems — comprising agents driven by internal energy and dissipative interactions — display even richer dynamics \cite{bowick2022symmetry}. These systems challenge traditional conservation laws as momentum and energy can be redistributed through environmental interactions \cite{ramaswamy2006mechanics,ramaswamy2010mechanics}, leading to non-trivial behaviors such as wall-accumulation in bacteria \cite{Sartori2018Wall} and spontaneous reorientation in cockroaches \cite{Li2015Terradynamically}. Other remarkable phenomena include the emergence of patterns and collective order driven by repulsive interactions, observed in systems such as molecular motors \cite{schaller2010polar, sumino2012large}, self-propelled rods \cite{bar2020self, shi2018self, chate2020dry}, and shaken grains \cite{deseigne2010collective, kudrolli2008swarming, kumar2014flocking}. 

\begin{figure}[t!]
\includegraphics[width=1.0\columnwidth]{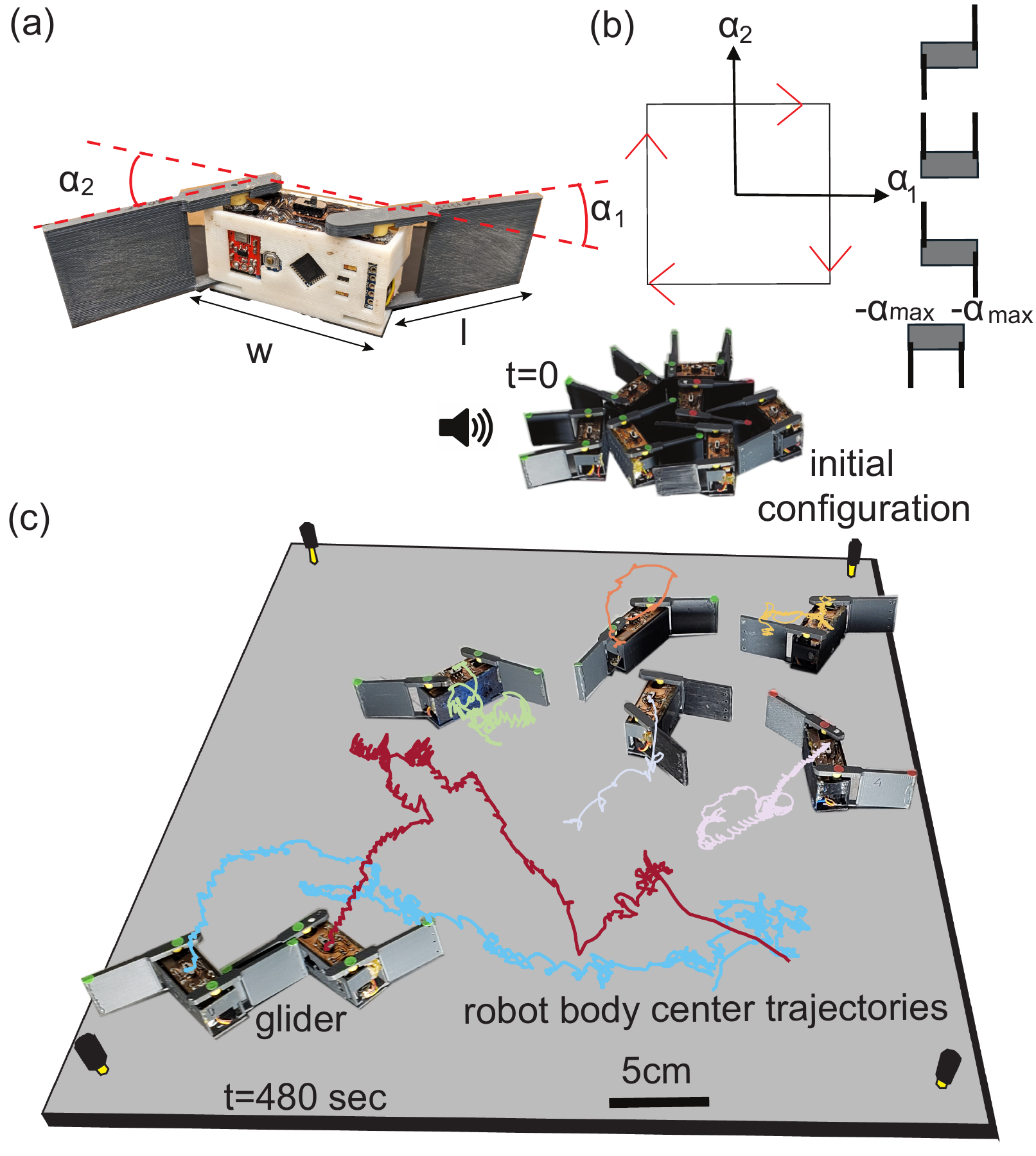}
\caption{(a) Configuration of a smart active particle (smarticle), parametrized by arm angles $\alpha_1$ and $\alpha_2$ relative to the body. The smarticle rests on its central body, with arms having slight ground clearance.  (b) Square gait configurations of a smarticle in the shape space defined by $\alpha_1,\alpha_2$ with amplitude $\alpha_{max}$. Red arrows indicate the direction of gait traversal.  (c) The emergence of a locomoting glider from a collective of seven smarticles, visualized through body center trajectories, with the final path traced at $t=480$ seconds. Top: compressed collective at $t=0$.}
\label{fig:Discovery}
\end{figure}

Active extended objects capable of self-deformation using internal energy exhibit distinctive locomotion and mechanical properties \cite{lauga2009hydrodynamics,lauga2011life,goldman2014colloquium, lauga2008no, marvi2014sidewinding,wagner2013crawling, hu2009mechanics,hatton2013geometric}, distinguishing them from traditionally studied active agents. While active agents reveal compelling behaviors in large numbers, shape-changing objects demonstrate unique emergent phenomena such as mechanical diffraction \cite{schiebel2019mechanical, rieser2019dynamics,zhang2021friction}, emergent locomotion \cite{brandenbourger2021active}, and lattice traversal \cite{wang2023mechanical}, even with fewer agents due to their high degrees of freedom. Collectively they exhibit phenomena like gait synchronization in nematode clusters and coordinated movement in undulatory robots \cite{yuan2014gait, quillen2021metachronal, peshkov2022synchronized, zhou2021collective, zhou2022lateral}, and form complex structures such as worm blobs and ant bridges \cite{ozkan2021collective, patil2023ultrafast, mlot2011fire}. Purcell-inspired three-link robots, including \emph{smarticles} (smart active particles) \cite{purcell1977life, savoie2015smarticles, savoie2018phototactic, savoie2019robot}, serve as a versatile platform for studying these systems. Though isolated smarticles exhibit limited motility, they demonstrate nontrivial dynamics through collisions, forming dynamically stable bound states that translate persistently despite noise and complex internal dynamics. By coupling shape-deformation with feedback mechanisms, these systems offer insights into emergent task-oriented behaviors \cite{manning2023essay,levine2023physics, vansaders2023informational, falk2021learning}.

Here, we study unconfined smarticle collections, where we observe that the individally immotile robots can emergently and dynamically bind in pairs to form physical ``gliders." These gliders ballistically travel several body lengths through repulsive collisional interactions. Experimentally validated simulations provide insights into glider formation and stability; these principles are applied to stabilize an otherwise unstable gliding mode through a contact sensor-based feedback strategy.

\begin{figure}[ht!]
  \includegraphics[width=0.9\columnwidth]{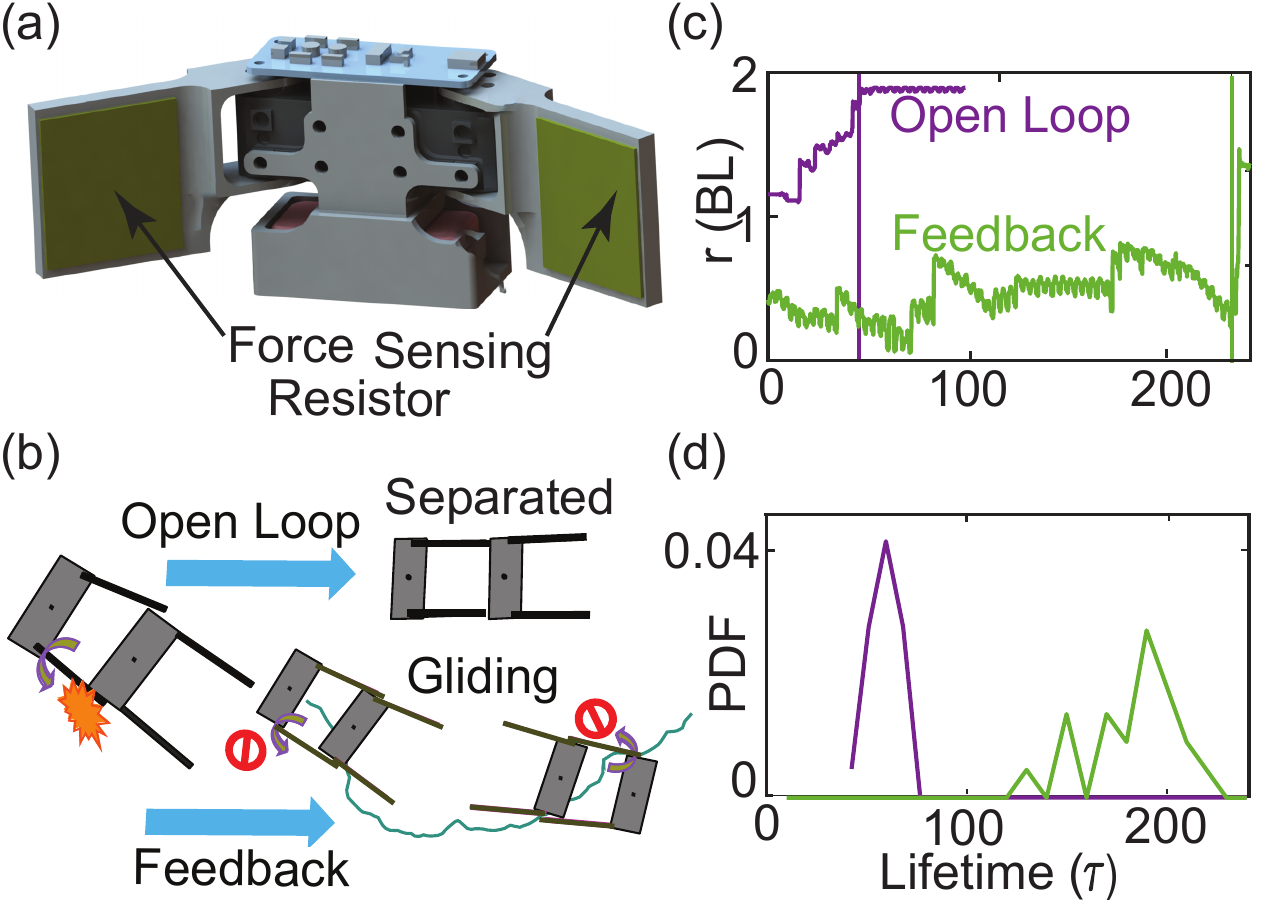}
\caption{
(a) Smarticles equipped with force sensing resistors on the four arm faces stabilize the sterically unstable C2 configuration (details in SI). (b) Mechanism illustrating the destabilization of open-loop and stabilization of feedback-enabled C2 gliders.
(c) Time evolution of $r$ for a representative open-loop and feedback-glider trial. Open-loop dyads push apart to stable fixed points, while feedback-enabled dyads remain in a transient mode for longer periods due to concavity modulation. Solid lines mark glider break-down.
(d) Glider lifetime distributions for open-loop and feedback-enabled trials.} 

  \label{fig:Feedback Stabilization}
\end{figure}

\emph{Experimental Apparatus.}
The smarticles, each with a mass of \emph{m} = 34.8 $\pm$ 0.6g, are studied under open-loop control. They consist of three links: two side arms, each having a length of \emph{l} = 5.0 cm and a thickness of 0.3 cm, and a central link (body) with a width of \emph{w} = 5.4 cm and a depth of \emph{D} = 2.2 cm Fig.~\ref{fig:Discovery}(a). 7 robots were placed on a $60 \times 60\, \mathrm{cm}^2$ aluminum plate, leveled to $\leq 0.1 ^{\circ}$. The gait -- the sequence of shape-changing motions -- of each smarticle (depicted in Fig.~\ref{fig:Discovery}(b)) was inspired by the dynamics of Purcell’s three-link swimmer \cite{purcell1977life,hatton2013geometric, Tam2007Purcell}. The periodic shape-changing motions are achieved by actuating the revolute joints connecting the central link to the side arms using a programmable servo motor, with a gait period of 1.6\ seconds. The side arms were rotated at the maximal motor speed until reaching the target arm rotation amplitude angle. 
The ground clearance of the arms prevents the inertial impulse of the actuated arms from being transmitted to the central body, which results in negligible motion of the central link. In addition, any motion induced by arm actuation is limited by the fiction between the central link and the underlying surface, contributing 0.0015 W (~75 $\mu$\text{m}) per cycle.  Therefore, individual smarticles do not move significant distances on their own. To enable contact feedback sensing, we designed a scaled-up version with a total mass of \emph{m} = 175 $\pm$ 0.5g, arm width and thickness of 6.0 cm and 0.7 cm, respectively, and body width and depth of 6.5 cm and 3.5 cm, respectively. We mounted force resistive sensors on all four arm faces to detect contacts based on impact thresholds (Fig.~\ref{fig:Feedback Stabilization}(a), Fig. S6).

\emph{Observation of Gliders.}
A collection of seven smarticles was initialized in a densely packed configuration, with all robots beginning their shape-changing motions simultaneously and in phase,triggered by a short audio pulse with a frequency of 1000 Hz, as shown in Fig.~\ref{fig:Discovery}(c). We expected the smarticles to push each other away through collisions, causing them to expand into a relaxed state, where no robots were in contact with each other. However, in 64\% of trials (97 out of 151), we observed that a pair of robots moved together for $\geq 100$ gait periods, traveling an average distance of $2.9 \pm 0.2$ $W$, as shown in Fig.~\ref{fig:Discovery}(c) (Supplementary Movie 1). 
The formation, stability, and locomotion of these dynamically bound pairs was surprising, given that the robots interacted locally repulsive collisions. The discrete and highly nonlinear collisions made theoretical analysis challenging. To systematically study the gliders, we developed a simulation based on the open-source physics engine Chrono (see SM for details) and calibrated it against experiments (see Fig. S1). In the following sections, we discuss the anatomy, binding mechanism, and stability of the gliding dyads. 


\begin{figure}[ht!]
\includegraphics[width=0.9\columnwidth]{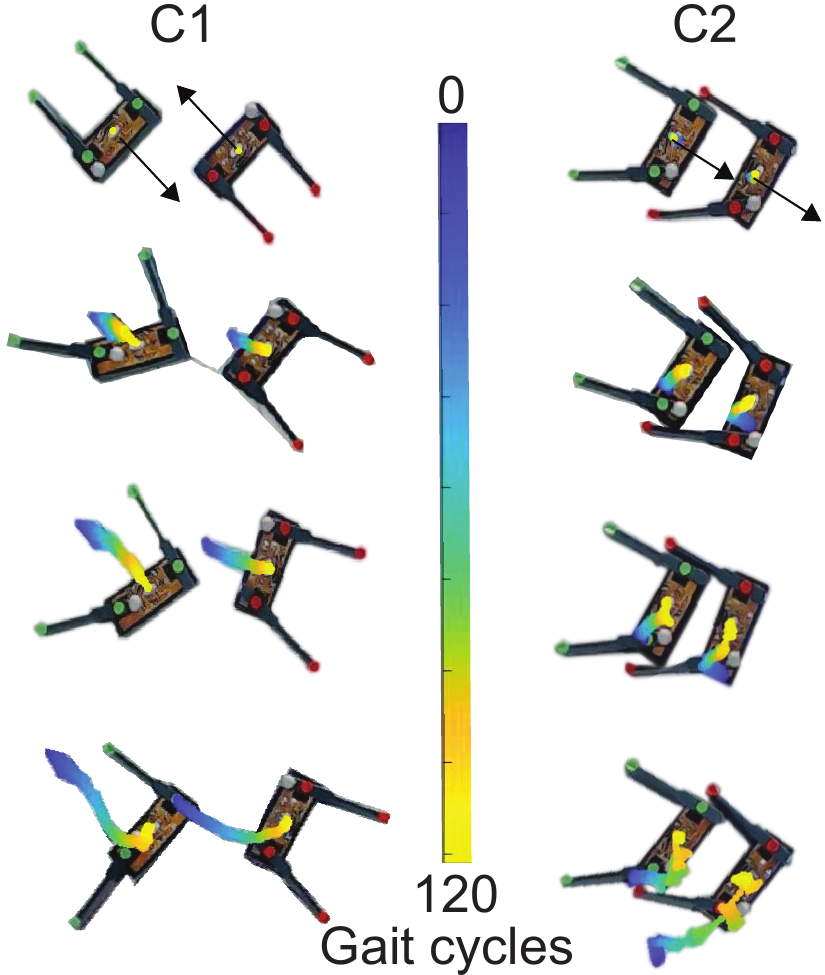}
\caption{ Strobed snapshots of the asymmetric emergent dyad configurations ($C1$ - almost anti-aligned, $C2$ - almost aligned), gliding as a bound entity over time. Arrows represent the body normal vectors.}
\label{fig:GliderModes_Visualization}
\end{figure}

\begin{figure}[ht!]
\includegraphics[width=0.9\columnwidth]{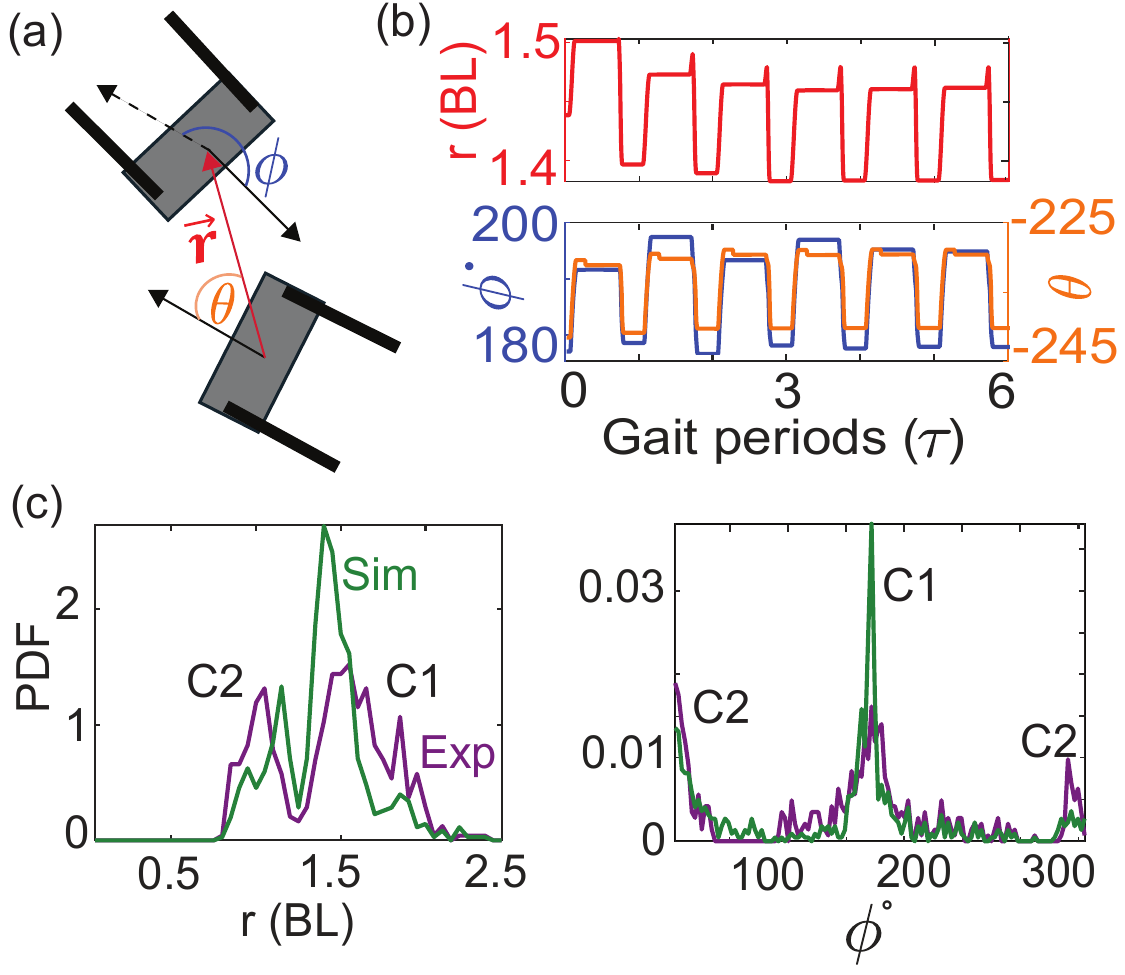}
\caption{(a) A gliding dyad is defined by a pair of smarticles with relative position ($\vec{r}$, $\theta$) in polar coordinates and relative body orientation ($\phi$) as the difference between their normal vector headings.
(b) Temporal evolution of $r$ in body lengths and the angular coordinates $\theta$, $\phi$ over six periods for a gliding dyad.
(c) Distributions of $r$ and $\phi$ for all emergent dyads in simulations and experiments, showing two peaks corresponding to modes C1 and C2.}
\label{fig:GliderModes_Data}
\end{figure}

\emph{Glider Configurations.}
Experiments and simulations (N = 151 trials) revealed that the gliders adopted two distinct configurations. These configurations exhibit a slight deviation from perfect anti-alignment and alignment, similar to chiral phases observed during non-reciprocal phase transitions \cite{fruchart2021non} in various non-equilibrium systems. We refer to them as: (i) ``$C1$," where the two smarticles are nearly anti-aligned, with their absolute orientations differing by approximately $180$ degrees, and (ii) ``$C2$," where the two smarticles are oriented in approximately the same direction. Snapshots of the glider configurations are shown in Fig.~\ref{fig:GliderModes_Visualization}. To characterize these distinct configurations, we measured their inter-smarticle distance $ r $, the polar angle $\theta$ and their relative orientation $\phi$, as defined in Fig.~\ref{fig:GliderModes_Data}(a).
The time evolution of the relative coordinates $r$, $\theta$, and $\phi$ over 6 periods for a glider trajectory (Fig.~\ref{fig:GliderModes_Data}(b)) demonstrates consistent behavior across many periods, indicating a limit cycle \cite{eldering2016role}. The distribution of the relative coordinates of the gliders observed in experiments and simulation shows distinct peaks in their histograms (Fig.~\ref{fig:GliderModes_Data}(c)(i-ii)), corresponding to the glider configurations $C1$ and $C2$ (Supplementary Movie 2). Furthermore, the two glider configurations exhibited distinct lifetimes (Fig. EM~\ref{fig:GliderLifetimes}), with $C1$ gliders persisting longer than $C2$ gliders, as shown by the long tail in the $C1$ lifetime distribution. For trials lasting 300 gait periods, some $C1$ gliders remained bound and locomoting until stopped. The phase drift in the gaits of glider-forming smarticle pairs during experiments caused slight variations in the distribution of $r$ and $\phi$, compared to simulations (Fig.~\ref{fig:GliderModes_Data}(c)(i-ii)).

\begin{figure}[ht!]
  \includegraphics[width = 1.0\columnwidth]{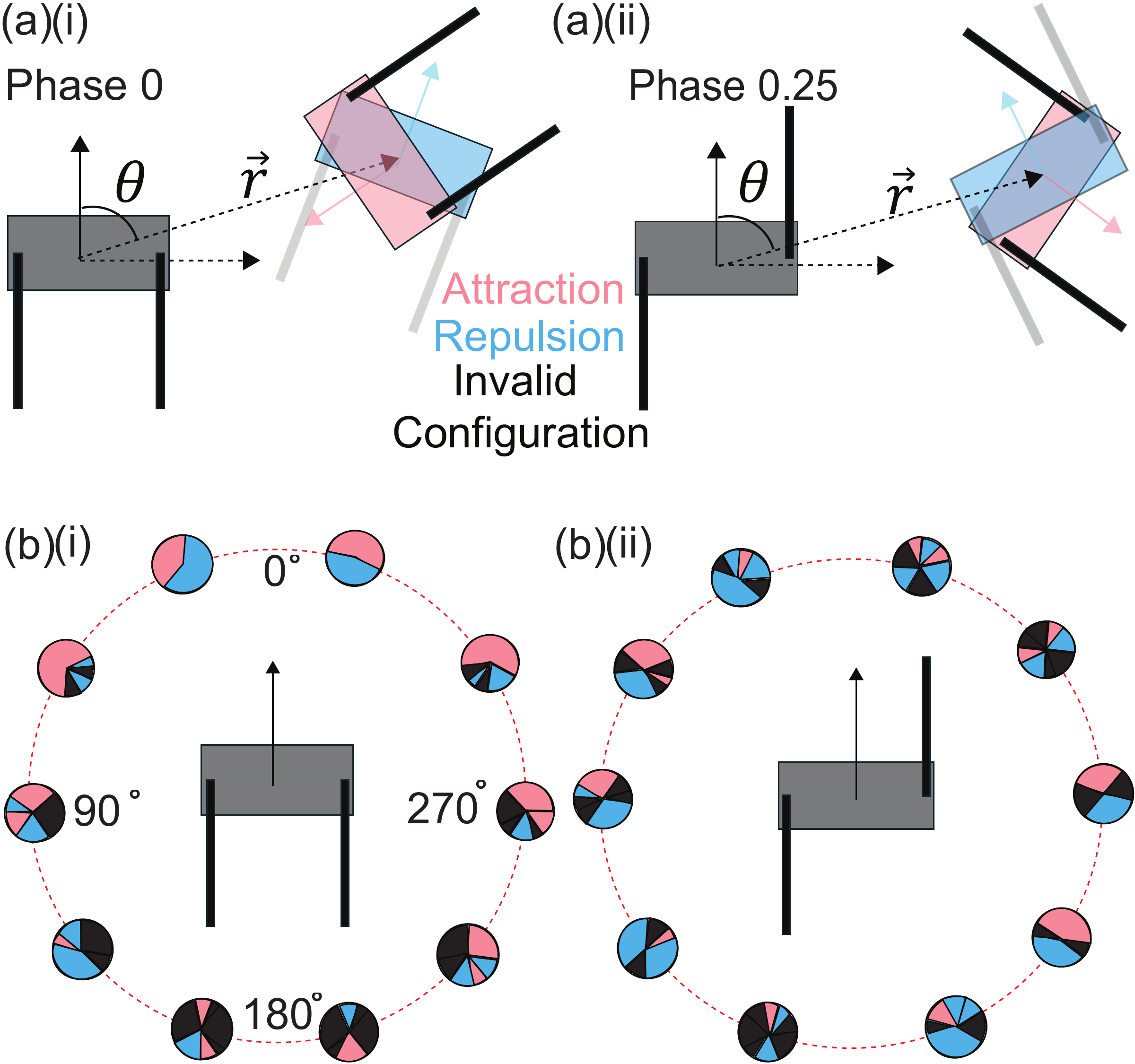}
\caption{(a) (i) Polar grid scan around a reference smarticle (gray) at a constant radius ($r = 1.3$ BL) for initial phase 0, showing an attracted (pink) and repelled (blue) initial configuration. The heading of the second smarticle is varied at a fixed polar angle, $\theta$.
(ii) The corresponding scan for phase 0.25 shows the change in the basin of attraction across different initial robot shape spaces.
(b) (i-ii) The scanned space is divided into regions of attraction (pink), repulsion (blue), and invalid configurations (black) for the two phase values.}
  \label{fig:Constant radius basin of attraction.}
\end{figure}

\emph{Binding Mechanism.}
Repulsive collisions between two convex-shaped bodies typically cause them to repel. However, smarticle pairs can remain bound as they develop concave shapes during their shape-changing motions. This enables the net contact impulse during a collision between the two robots to produce an attractive effect, causing them to move together at least once during the gait. Friction between the smarticle's base and the underlying surface prevents coasting, while periodic attractive contact forces during collisions keep them bound over multiple gait periods. However, under certain conditions —- dependent on the smarticle pairs' relative positions, orientations, and gait phase -— the pairs can harness attractive collisions that keep them bound.
To identify the binding conditions, we simulated collisions by fixing one robot (gray) and varying the absolute heading of the second robot across a constant-radius polar grid ($r = 1.3$ BL, $\theta, \phi$) for two gait phases, $p= 0, 0.25$, as shown in Fig.~\ref{fig:Constant radius basin of attraction.}(a). Once initialized at the same phase, we tracked the lifetimes of bound pairs over 75 gait periods and partitioned the relative configuration space into regions of attraction (pink) and repulsion (blue), based on the outcome of each robot pair. The attracted configurations correspond to initial conditions that formed stable glider and survived the entire trial. These configurations converge to the periodic deformation pattern characteristic of the attractor (Fig.~\ref{fig:GliderModes_Data}(b)) within a few periods. Moreover, robot pairs' binding status can be determined from their configuration after the first gait period, where their relative orientations dictate whether the subsequent collision results in attraction or repulsion.

The results indicate that the formation of a bound pair between two smarticles is governed by a non-reciprocal binding affinity, which depends on their polar angle $\theta$ and relative headings $\phi$. Robot pairs with initial configurations in the first and fourth quadrants (with respect to $\theta$) of the interaction range are likely to form bound pairs (pink sectors) when their normal vectors are antiparallel. Conversely, robot pairs with parallel normal vectors are likely to repel (blue sectors) or be physically inadmissible (black sectors).  

\begin{figure}[ht!]
  \includegraphics[width = 1.0\columnwidth]{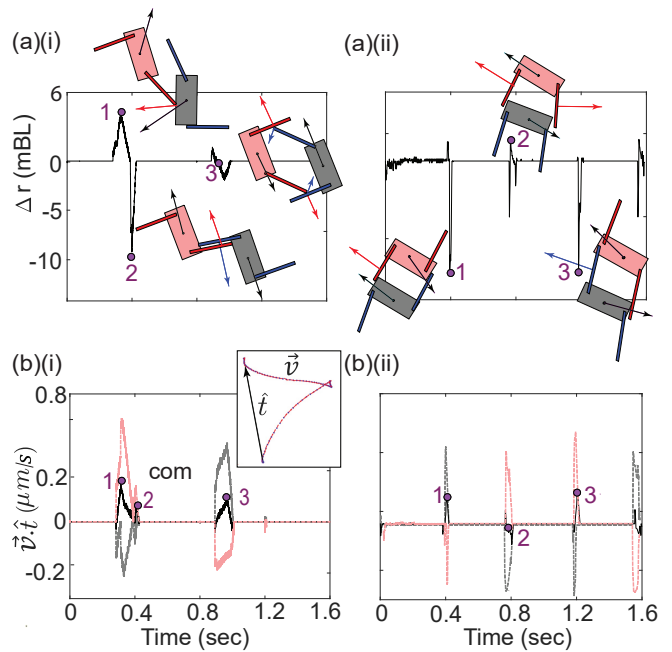}
\caption{(a) Collision events for a C1 dyad (i), where attraction arises from arm hooking and repulsion from arm-body collisions, and for a C2 dyad (ii), where attraction results from alternate arms colliding with each other's bodies and repulsion from steric interactions when one robot's body is fully enveloped by the arms of the other. The change in robot separation, $\Delta r$, over time shows sharp spikes that correlate with attraction events for both dyads.
(b) Projections of the robots' instantaneous velocities onto the effective stepping vector reveal individual displacements along the transport direction, highlighting a leader-follower dynamic where one robot leads and the other is carried along. This links transport to repulsive collisions in both dyad types (i-ii).} 
  \label{fig:Binding and Transport Mechanism.}
\end{figure}

In contrast, configurations in the second and third quadrants are generally repelled or physically inadmissible, as shown in Fig.~\ref{fig:Constant radius basin of attraction.}(b)(i-ii) and Figs. S4 and S5. Once on the attractor, we analyzed a single gait period for the two glider configurations to identify collisions driving attraction and transport during the cycle. Plotting the instantaneous change in robot separation, $\Delta r = r_i - r_{i-1}$ (where $r_i$ is the separation at time $t_i$ during the gait period $T$), over time reveals that for the C1 dyad, attraction arises when the arms briefly hook onto each other during the cycle (Fig.~\ref{fig:Binding and Transport Mechanism.}(a)(i), collision 2). In contrast, Fig.~\ref{fig:Binding and Transport Mechanism.}(a)(ii) reveals that for the C2 dyad, attraction results from the robots' alternating arms bracing each other's bodies (collisions 1 and 3). Both events are marked by a sharp decrease in $\Delta r$, as seen in Fig.~\ref{fig:Binding and Transport Mechanism.}(b)(i-ii). 

The net displacement of the paired robots in the heading direction during a single cycle can be calculated by taking the area under the curve of the dot product of their instantaneous velocity $\vec{v}$ and the glider's effective stepping vector of the center of mass \( \hat{t} \) for the cycle. For the C1 dyad, transport occurs primarily when the arm of the robot maintaining the orientational asymmetry (Leader) briefly collides with the body of the other robot (Follower) before hooking onto its arm to create attraction. This collision 1, shown in panel Fig.~\ref{fig:Binding and Transport Mechanism.}(a)(i), causes the robots to move in opposite directions, with the leading robot driving the net transport (Fig.~\ref{fig:Binding and Transport Mechanism.}(c)(i)). In contrast, the C2 dyad exhibits transport driven by collisions, where one robot's arms alternatively envelope the body of the other every half-cycle, as shown in collision 2 of Fig.~\ref{fig:Binding and Transport Mechanism.}(a)(ii). These collisions cause both robots to move in opposite directions (Fig.~\ref{fig:Binding and Transport Mechanism.}(c)(ii)). However, the displacement magnitudes for both robots are almost similar, and the effect accumulates over multiple cycles, eventually causing the dyad to break apart.  

To test our understanding of the stabilized binding and gliding, we conclude by demonstrating how a simple control strategy extends the lifetime of the short-lived C2 glider.

\emph{Feedback stabilization using tactile sensing.} The slightly asymmetric C2 glider typically unbound permanently (and thus stopped translating) within 60-70 cycles. This glider conformation involved the two robots fully enclosing each other in the space between their arms every half cycle, as shown in collision 2 of Fig.~\ref{fig:Binding and Transport Mechanism.}(a)(ii), causing them to move further apart due to compression. These steric effects, accumulated over multiple cycles, caused the bound pair to eventually disengage and remain barely in contact. To mitigate this repulsion, we employed a simple contact-sensing feedback strategy. We constructed slightly larger robots capable of housing  resistance-based force sensors mounted on all four faces of the arms, see Fig.~\ref{fig:Feedback Stabilization}(a). We calibrated these force sensors based on the forces generated during the compression collision in the typical C2 conformation (Fig. S6). Using this impact threshold, we implemented a feedback strategy where the robots stopped moving their arms upon detecting an impact within that range and resumed movement from that position in their next gait cycle, as shown in Fig.~\ref{fig:Feedback Stabilization}(b).

The impact-based amplitude modulation stabilized the C2 glider. Fig.~\ref{fig:Feedback Stabilization}(c) shows a plot of $r$ vs. time (in gait periods) for a representative open-loop and feedback-controlled C2 glider, illustrating how the closed-loop strategy keeps the value of $r$ fluctuating transiently. The distribution of lifetimes for C2 gliders with different open-loop arm amplitudes (with $70^{\circ}$ as the cutoff, beyond which no C2 gliders form) compares to the feedback-stabilized glider in Fig.~\ref{fig:Feedback Stabilization}(d). The control strategy significantly improved the lifetime of this unstable open-loop excitation,  keeping it dynamically evolving, as shown in the relative configuration space in Fig. S7. The transport properties of the various C2 gliders with different arm amplitudes are shown in Fig. S8. The open-loop C2 glider with an amplitude of $85^{\circ}$ and the feedback-controlled C2 glider move almost ballistically, but the longer lifetime of the feedback-stabilized glider ensures it to transport over longer distances, as shown in Supplementary Movie 4.


\emph{Conclusion.}
We studied ensembles of shape-changing robots (smarticles) in experiments and simulations, revealing the novel phenomenon of \emph{gliders}-- dynamically bound pairs that travel together with almost ballistic locomotion, over many body lengths and enduring hundreds of rigid-body collisions between them. We characterize gliders by their relative spacing and orientation, demonstrating that their formation
emerges from a delicate interplay between the robot's geometric positioning and arm phasing. The stability and lifetime of these gliders depends on how the robots collide and envelop each other within the space between their arms during a gait cycle. Using an impact-sensing feedback strategy to modulate the concavity, we significantly enhanced the binding of the unstable mode. Surprisingly, these physical gliders can assemble into long structures upon colliding with other robots, giving rise to long-range order in the collective, similar to non-reciprocally interacting spin models \cite{loos2023long,dadhichi2020nonmutual} (Supplementary Movie 5). We also observed a transition from static bound states to locomoting gliders, depending on the amount of non-reciprocity programmed into the gaits. A detailed analysis of how glider transport can be understood and modulated by enumerating the symmetries of glider configurations and the gaits executed by the robots is in preparation\cite{Vardhan2025GliderTransport}. We suggest that gliders may emerge in diverse active matter systems, where interacting shape-changing objects adopt concave shapes during non-reciprocal deformations in a sufficiently damped environments, highlighting the rich dynamics of highly deformable active bodies. These findings highlight principles for dynamically entangling collectives, paving the way for directed self-assembly and transport in active shape-changing systems.

\medskip 

\emph{Acknowledgments}
We thank Will Savoie for the discovery of gliders, and Pavel Chvykov, Zachary Jackson, and Aryaman Jha for stimulating discussions. We also appreciate the feedback from the organizers and participants of the \textit{Active Matter in Complex and Crowded Environments} workshop. We acknowledge the use of ChatGPT in refining this manuscript. This work was supported by ARO MURI Grant W911NF-13-1-034, the School of Physics at the Georgia Institute of Technology, the QBIOS PhD program, and a Dunn Family Professorship (D.I.G).


%

\section*{End matter}

\emph{Glider Lifetimes.}

The C1 glider, or nearly antisymmetric mode, remained sterically stable due to sustained periodic attraction from the hooking interaction between its arms (Fig.~\ref{fig:Binding and Transport Mechanism.}(a)(i)).

\begin{figure}[ht!]
   \centering
   \includegraphics[width=0.8\columnwidth]{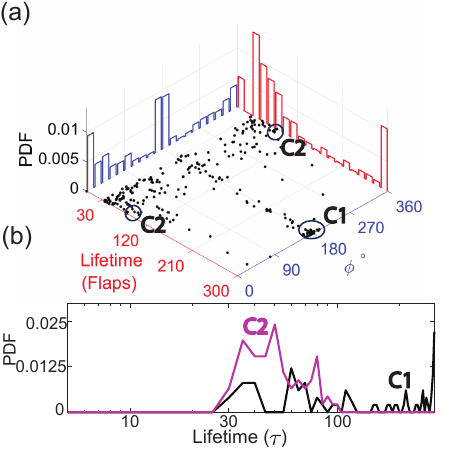} 
   \caption{
   (a) Lifetime of emergent dyads vs. relative angle $\phi$, showing the lifetimes of the two configurations. The scatter plot shows that the C1 configuration is more stable dynamically bound mode, with lifetimes spanning the entire trial. Histograms display the individual distributions for lifetime and relative orientation $\phi$.
   (b) Lifetimes of the two configurations on a semi-log scale, illustrating the presence of long-lived C1 gliders, denoted by the peak around 300 flaps.
   }
   \label{fig:GliderLifetimes}
   \end{figure}

 In contrast, the C2 glider disintegrated under cyclic compression interactions (Fig.~\ref{fig:Binding and Transport Mechanism.}(a)(ii)). Lifetimes of emergent gliders exceeding 30 periods -— the time for the expanding cloud to reach steady state -— plotted against relative orientation \( \phi \) show that C1 gliders, clustered around \( \phi = 180^\circ \), persist throughout the trial (Fig.~\ref{fig:GliderLifetimes}(a)). In contrast, C2 gliders, characterized by symmetric clusters at \( \phi = 0, 360^\circ \), break apart faster, lasting only 70–100 cycles. To highlight these effects, we plotted lifetimes on a semi-log scale (Fig.~\ref{fig:GliderLifetimes}(b)), where the sharp peak at 300 indicates that the surviving gliders at the end of the trial remained in the C1 configuration.

\emph{ Glider Robustness and Transport.}

To test the robustness of bound-pair states and their subsequent locomotion, we systematically varied the arm amplitude (maximum rotation angle $ \alpha_{max} $) while keeping the rotational speed constant. We then studied the persistence of bound pairs for shapes with different concavities \cite{rosenfeld1985measuring}. We performed experiments by fixing one smarticle and initializing the other at 21 points on a polar grid of radius 1 BL, across a range of $ \alpha_{max} $. We observed different types of bound states: some in which the smarticles maintained periodic contact over multiple cycles, and others where they disengaged for a few cycles before re-establishing contact.

\begin{figure}[ht!]
  \includegraphics[width = 0.9\columnwidth]{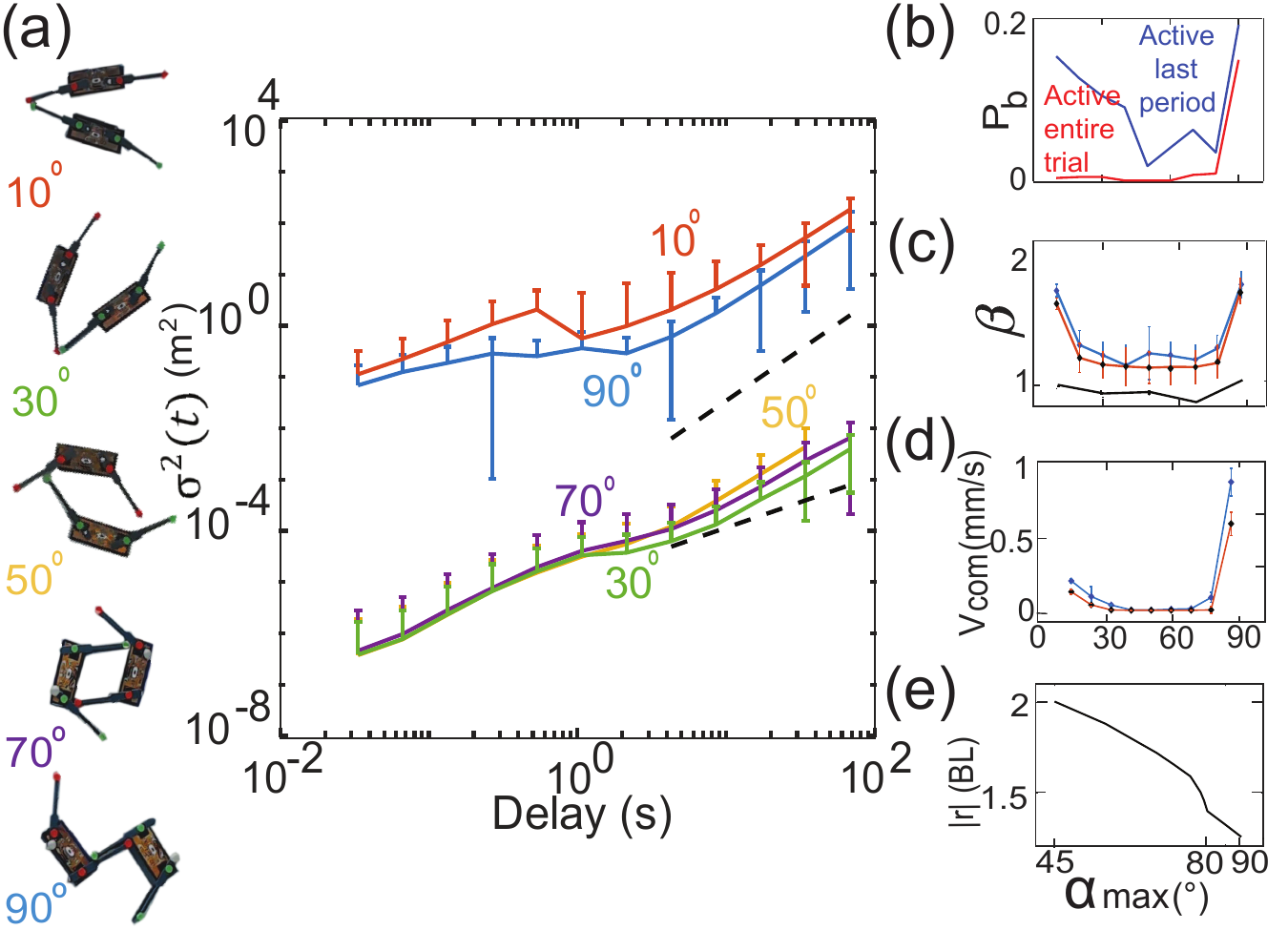}
  \centering
  \caption{
  (a) MSD $ \langle \sigma^2(t) \rangle $ vs. time delay for five arm amplitudes, with snapshots of the corresponding bound pairs. 
  (b) Binding probability $ P_b $ vs. arm amplitude for random initial conditions; blue: pairs active at trial end, red: pairs active throughout the trial. 
  (c) MSD exponent $ \beta $ vs. $ \alpha_{\max} $ for bound pairs; blue (experiment) and orange (simulation) curves with standard deviation bars. Black: exponent for time-symmetric gait in simulations. 
  (d) Bound pair center-of-mass speed $ V_{com} $ vs. $ \alpha_{\max} $. 
  (e) Steady-state separation $ r $ vs. $ \alpha_{\max} > 45^\circ $ from simulations of initially bound smarticles.}
  \label{fig:BoundPairTransport}
\end{figure}

The mean-square displacement (MSD) vs. time for glider trajectories surviving more than 30 cycles is shown in Fig.~\ref{fig:BoundPairTransport}(a) for five different arm amplitudes. The bound pairs for $90^{\circ}$ and $10^{\circ}$ move nearly ballistically, with $ \langle {\sigma } ^ 2(t) \rangle  \propto t ^ {\beta} $  (where MSD $ \langle {\sigma } ^ 2(t) \rangle = \langle \vec{x}^2(t) \rangle -{\langle \vec{x}(t) \rangle}^2  $ and exponent $\beta \approx 1.8$), 
For intermediate arm amplitudes, the motion is almost diffusive. We observed that gliders in experiments exhibit a higher $ \beta $ and center-of-mass speed $ V_{com} $ than those in simulations, as shown in Fig.~\ref{fig:BoundPairTransport}. We posit that experimental noise from the motors, collisions, and friction with the floor enhances the transport of bound pairs with intermediate arm amplitudes ($40^{\circ} - 60^{\circ}$), but the pairs bind and unbind frequently. The lack of stability for intermediate $ \alpha_{max} $ correlates with a decrease in binding probability (Fig.~\ref{fig:BoundPairTransport}(b)). 

Bound pair speed and transport decreased sharply as $ \alpha_{max} $ decreased from $90^{\circ}$, but surprisingly increased again in the low-amplitude regime (see $10^{\circ} - 20^{\circ}$ range in Fig.~\ref{fig:BoundPairTransport}(c-d)). We attribute this anomaly to inertial effects, which enhance the drift of a single smarticle, as shown in Fig. S2. Since the smarticle arms were driven at a constant motor speed, at low rotation angles, the arm rotation time matched the overlap time between the motions of two arms, as shown in Fig. S3. This significantly increased the motility of individual smarticles. As a result, dyad pairs that remained bound at low $ \alpha_{max} $ also showed improved gliding speed, as shown in Fig.~\ref{fig:BoundPairTransport}(c-d). 
To probe how arm amplitude affects glider locomotion, we simulated smarticles pairs in a bound formation, performing gaits with varying $\alpha_{max}$ values greater than $45^{\circ}$. A clear trend appears, where $r$ monotonically decreases as $\alpha_{max}$ increases (Fig.~\ref{fig:BoundPairTransport}(e)), with a rapid drop in $r$ around $85^{\circ}$. Coincidentally, this is also roughly the amplitude at which the $\beta $ and $ V_{com} $ increase sharply (Fig. ~\ref{fig:BoundPairTransport}(c-d)). Simulations revealed that for $\alpha_{max}$ $\leq$ $85^{\circ}$, the smarticles interact with only one arm, while above this threshold, they experience an additional collision with the other arm. The additional collision brings the smarticles closer together, leading to a third collision between an arm and the central link. Since super-diffusive transport becomes effective for $\alpha_{max}$ $\geq$ $85^{\circ}$, we attribute most of the locomotion to these additional collisions.

\end{document}